\documentclass[
twocolumn,
]{ceurart}

\usepackage{listings}
\usepackage{natbib}
\usepackage{color, colortbl}
\definecolor{LightCyan}{rgb}{0.88,1,1}
\definecolor{LightGreen}{rgb}{0.88,1,0.88}
\definecolor{LightRed}{rgb}{1,0.88,0.88}

\begin{document}

\copyrightyear{2024}
\copyrightclause{Copyright for this paper by its authors.
  Use permitted under Creative Commons License Attribution 4.0
  International (CC BY 4.0).}

\conference{CLiC-it 2024: Tenth Italian Conference on Computational Linguistics, Dec 04 — 06, 2024, Pisa, Italy}

\title{SLIMER-IT: Zero-Shot NER on Italian Language}

\author[1,2]{Andrew Zamai}[
email=andrew.zamai@unisi.it
]
\address[1]{Università degli Studi di Siena, Italy}
\address[2]{expert.ai, Siena, Italy}

\author[2]{Leonardo Rigutini}
[email=lrigutini@expert.ai]

\author[1]{Marco Maggini}
[email=marco.maggini@unisi.it]

\author[2]{Andrea Zugarini}
[email=azugarini@expert.ai]
\cormark[1]

\newcommand{\az}[1]{\color{black} #1}
\newcommand{\andrew}[1]{\color{black} #1}

\cortext[1]{Corresponding author.}

\begin{abstract}
Traditional approaches to Named Entity Recognition (NER) frame the task into a BIO sequence labeling problem. Although these systems often excel in the downstream task at hand, they require extensive annotated data and struggle to generalize to out-of-distribution input domains and unseen entity types.
On the contrary, Large Language Models (LLMs) have demonstrated strong zero-shot capabilities. 
While several works address Zero-Shot NER in English, little has been done in other languages. In this paper, we define an evaluation framework for Zero-Shot NER, applying it to the Italian language. Furthermore, we introduce SLIMER-IT, the Italian version of SLIMER, an instruction-tuning approach for zero-shot NER leveraging prompts enriched with definition and guidelines. Comparisons with other state-of-the-art models, demonstrate the superiority of SLIMER-IT on never-seen-before entity tags.
\end{abstract}

\begin{keywords}
  Named Entity Recognition \sep
  Zero-Shot NER \sep
  Large Language Models \sep
  Instruction tuning
\end{keywords}

\maketitle

\section{Introduction}
Named Entity Recognition (NER) plays a fundamental role in Natural Language Processing (NLP), often being a key component in information extraction pipelines. 
The task involves identifying and categorizing entities in a given text according to a predefined set of labels.
While \textit{person}, \textit{organization}, and \textit{location} are the most common, applications of NER in certain fields may require the identification of domain-specific entities. 

\begin{figure}
  \centering
  \includegraphics[width=0.85\linewidth]{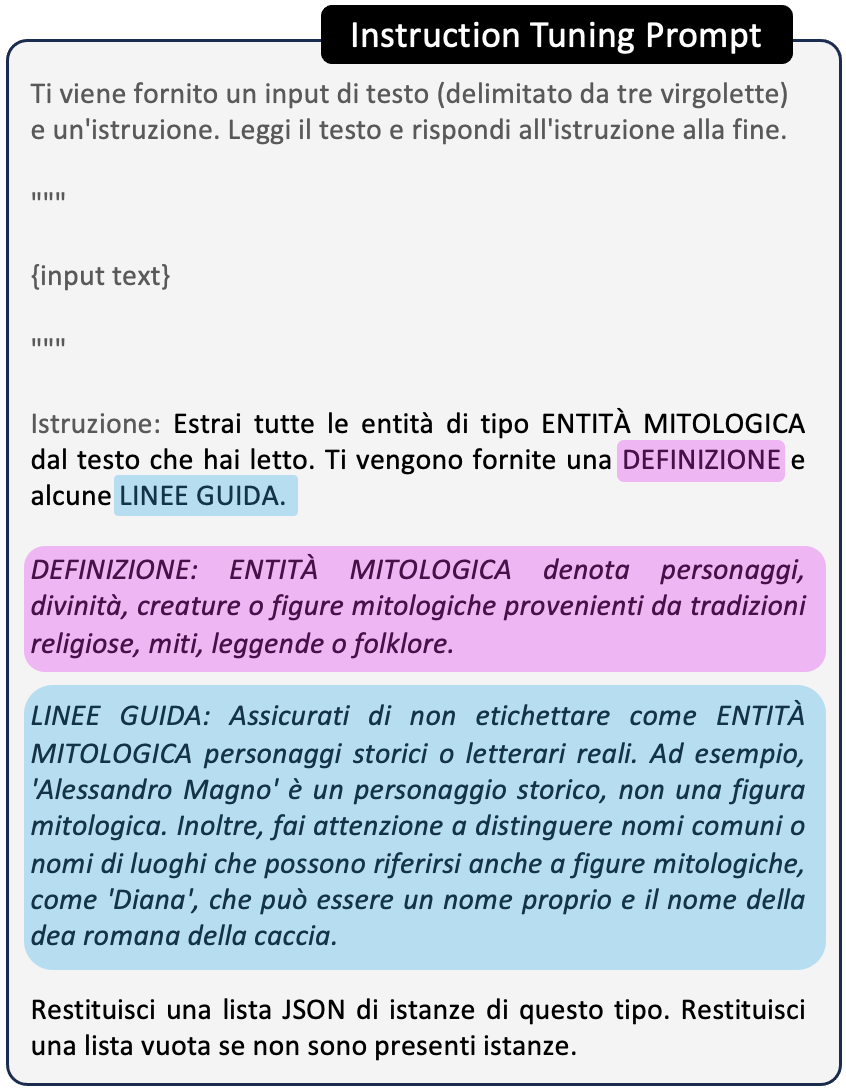}
  \caption{SLIMER-IT instruction tuning prompt. Dedicated entity \textit{definition} and \textit{guidelines} steer the model labelling.}
  \label{fig:SLIMER-IT prompt}
\end{figure}

Manually annotated data has always been critical for the training of NER systems \cite{Aprosio2023NERMuDAE}. Traditional methods tackle NER as a token classification problem, where models are specialized on a narrow domain and a pre-defined labels set \cite{li2020survey}. While achieving strong performance for the data distribution they were trained on, they require extensive human annotations relative to the downstream task at hand. Additionally, they lack generalization capabilities when it comes to addressing out-of-distribution input domains and/or unseen labels \cite{Aprosio2023NERMuDAE, sainz2024gollie, zhou2023universalner}. 

On the contrary, Large Language Models (LLMs) have recently demonstrated strong zero-shot capabilities. Models like GPT-3 can tackle NER via In-Context Learning \cite{radford2019language, brown2020language}, with Instruction-Tuning further improving performance \cite{wei2022finetuned, chung2022scaling, wang-etal-2022-super}. To this end, several models have been proposed to tackle zero-shot NER~\citep{wang2023instructuie,zhou2023universalner,sainz2024gollie,zaratiana2023gliner,ding2024rethinking, SLIMER}. In particular, SLIMER~\cite{SLIMER} proved to be particularly effective on unseen named entity types, by leveraging definitions and guidelines to steer the model generation.

However, little has been done for zero-shot NER in non-English data.  More in general, as pointed out in \cite{Aprosio2023NERMuDAE}, NER is understudied in languages like Italian, especially outside the traditional news domain and \textit{person}, \textit{location}, \textit{organization} classes. 

To this end, we propose in this paper an evaluation framework for Zero-Shot NER, and we apply it to the Italian language. 
In addition, we fine-tune a version of SLIMER for Italian, which we call SLIMER-IT\footnote{\url{https://github.com/andrewzamai/SLIMER_IT}}. In the experiments, we explore different LLM backbones and we assess the impact of Definition and Guidelines (D\&G). When comparing SLIMER-IT with state-of-the-art approaches, either using models pre-trained on English or adapted for Italian, results demonstrate SLIMER-IT superiority in labelling unseen entity tags.

\section{Related Work}
\label{sec:related_works}
Several works tackle Zero-Shot NER on English, such as InstructUIE \cite{wang2023instructuie}, UniNER \cite{zhou2023universalner}, GoLLIE \cite{sainz2024gollie}, GLiNER \cite{zaratiana2023gliner}, GNER \cite{ding2024rethinking} and SLIMER \cite{SLIMER}.
Most of them are based on the instruction tuning of an LLM and mainly differ in the prompt and output format design. GLiNER distinguishes itself by being a smaller encoder-only model, combined with a span classifier head, that achieves competitive performance at a lower computational cost.

As highlighted in SLIMER~\cite{SLIMER}, most approaches mainly focus on zero-shot NER in Out-Of-Distribution input domains (OOD), since they are typically fine-tuned on an extensive number of entity classes highly or completely overlapping between training and test sets.
In view of this, we proposed a lighter instruction-tuning methodology for LLMs, training on data overlapping in lesser degree with the test sets, while steering the model annotation process with a definition and guidelines for the NE category to be annotated. From this, the name SLIMER: Show Less, Instruct More Entity Recognition.

Although the authors of GLiNER propose also a multi-lingual model and evaluate zero-shot generalizability across different languages, neither they nor any other work has addressed the task of Zero-Shot NER specifically for the Italian language.

\paragraph{NER for Italian.} 
While NER has been extensively studied on English, less has been done in other languages, particularly outside the traditional general-purpose domains and entity labels set~\cite{MARRERO2013482}. Indeed, in Italian, most NER datasets focus on news and, more recently, social media contents~\cite{ICAB, EVALITA12, EVALITA16}. 
Currently, there has been no research into zero-shot NER, only a few exploratory studies into multi-domain NER. 
This challenge was introduced in the NERMuD task (NER Multi-Domain) at EVALITA 2023\footnote{\url{https://www.evalita.it/campaigns/evalita-2023/tasks/}}, in which one sub-task required to develop a single model capable of classifying the common entities - \textit{person}, \textit{organization}, \textit{location} - from different types of text, including news, fiction and political speeches. ExtremITA team~\cite{ExtremITA} addressed the challenge proposing the adoption of a single LLM capable of tackling all the different tasks at EVALITA 2023, among which NERMuD. All the tasks were converted into text-to-text problems and two LLMs (LLaMA and T5 based) were instruction-tuned on the union of all the available datasets for the challenge. 

\section{Zero-Shot NER Framework}\label{sec:zero-shot-framework}
In traditional Machine-Learning theory, a model $f$, trained for a task (e.g. NER) represented by a dataset $\mathcal{X},\mathcal{Y}$, 
is typically evaluated on an held-out test set sampled from the same task and distribution of the training. In zero-shot learning instead, a model is expected to go beyond what experienced during training. 
There are different levels of generalization indicating up to what extent the model goes beyond what directly learnt.

In the case of zero-shot NER, a model should be able to extract entities from inputs belonging to the same domain it was trained on (\textbf{in-domain}) and across other domains not encountered before (\textbf{out-of-domain}). Moreover, it should also generalize well to novel entity classes (\textbf{unseen named entities}).
In our zero-shot evaluation framework we aim to measure each level independently. Hence, we define an evaluation benchmark that includes a collection of NER datasets divided by degree of generalization.
In the following we describe the required properties to fit in. 






\paragraph{In-domain.} This evaluation helps measure how well the model can generalize from its training data to similar, but not identical, data. The model is evaluated on the same input-domains and named entities as those in the training set. This data often consists in the test partitions associated with each training set used for fine-tuning the model.

\paragraph{Out-Of-Domain (OOD).} OOD evaluation tests the model's ability to generalize to input texts from domains that it has not encountered during training. While the named entities have been seen during training, this type of evaluation is particularly challenging because different input domains often exhibit unique linguistic patterns and domain-specific terminology. 

\paragraph{Unseen Named Entities.} This evaluation tests the model's ability to identify and classify entities that has not encountered during its training phase. The tag set comprises fine-grained categories which are often specifically defined for the domain in which NER is deployed. Because of this, the input data may often be also Out-Of-Domain (OOD), making this evaluation include the previously mentioned OOD scenario as well. 


\section{SLIMER-IT}

To adapt SLIMER for Italian, we translate the instruction-tuning prompt of \cite{SLIMER}, as shown in Figure~\ref{fig:SLIMER-IT prompt}. The prompt is designed to extract the occurrences of one entity type per call. While this has the drawback of requiring $|\mbox{NE}|$ inference calls on each input text, it allows the model to better focus on a single NE type at a time. 

As in \cite{SLIMER}, we query gpt-3.5-turbo-1106 via OpenAI's Chat-GPT APIs to automatically generate definition and guidelines for each needed entity tag. The definition for a NE is meant to be a short sentence describing the tag. The guidelines instead provide annotation instructions to align the model's labelling with the desired annotation scheme. Guidelines can be used to prevent the model from labelling certain edge cases or to provide examples of such NE. Such an informative prompt is extremely valuable when dealing with unfamiliar entity tags, and can also be used to distinguish between polysemous categories. 

Finally, the model is requested to generate the named entities in a parsable JSON format containing the list of NEs extracted for the given tag.

\section{Experiments}
Experiments aim to assess our approach in Italian. We study the impact of guidelines and the usage of different backbones. Then, we compare our approach against state-of-the-art alternatives. 

\subsection{Datasets}
We construct the zero-shot NER framework (described in Section~\ref{sec:zero-shot-framework}) for Italian upon NerMuD shared task and Multinerd dataset. In particular, we use NerMuD to build in-domain and OOD evaluation sets, while Multinerd-IT is used to assess the behaviour in the unseen named entites scenario.

\paragraph{NERMuD.} NERMuD~\cite{Aprosio2023NERMuDAE} is a shared task organized at evalita-2023, built based on the Kessler Italian Named-entities Dataset (KIND)~\cite{KIND}. It contains annotations for the three classic NER tags: \textit{person}, \textit{organization} and \textit{location}.
Examples are organized in three distinct domains: news, literature and political discourses.
Unlike NERMuD, we restrict fine-tuning to a single domain. In such a way, we can evaluate both in-domain and out-of-domain capabilities of the model. In particular, we designate WikiNews (WN) sub-set for training and in-domain evaluation, being the most generic domain, while Fiction (FIC) and Alcide De Gasperi (ADG) splits are kept for out-of-domain evaluation only. 

\paragraph{Multinerd-IT.} To construct the unseen NEs evaluation set, we exploit Multinerd\footnote{\url{https://github.com/Babelscape/multinerd}}~\cite{multinerd}, a multilingual NER dataset made of 15 tags: \textit{person}, \textit{organization}, \textit{location}, \textit{animal}, \textit{biological entity}, \textit{celestial body}, \textit{disease}, \textit{event}, \textit{food}, \textit{instrument}, \textit{media}, \textit{plant}, \textit{mythological entity}, \textit{time} and \textit{vehicle}. We keep the Italian examples only. 
Such a dataset constitutes a perfect choice to assess models' capabilities on unseen NEs. Indeed, data belongs to the same news domain of the NERMuD split chosen for fine-tuning, but it includes a broader label set. Since we want to measure performance on never-seen-before entities, we exclude entity types seen in training, i.e. \textit{person}, \textit{organization} and \textit{location}. 
We also remove \textit{biological entity}, being poorly underrepresented, with a support of just 4 instances.

\subsection{Backbones}
We implemented several version of SLIMER-IT based on different backbone models. 
We consider similarly sized LLMs, all in the 7B parameters range. 
In particular, we selected five backbones: Camoscio\footnote{\url{https://huggingface.co/teelinsan/camoscio-7b-llama}}~\cite{camoscio}, LLaMA-2-7b-chat~\cite{touvron2023llama}, Mistral-7B-Instruct~\cite{jiang2023mistral7b}, LLaMA-3-8B-Instruct, LLaMAntino-3-ANITA-8B-Inst-DPO-ITA\footnote{\url{https://huggingface.co/swap-uniba/LLaMAntino-3-ANITA-8B-Inst-DPO-ITA}}~\cite{llamantino3anitapaper}.

LLaMA-2-7b-chat was originally used in SLIMER~\cite{SLIMER}, and LLaMA-3-8B-Instruct is the newest, improved version of it. As LLaMA family, Mistral-7B-Instruct is a multilingual model mainly English-oriented, but it has demonstrated greater fluency on Italian. 
Camoscio and LLaMAntino-3-ANITA-8B-Inst-DPO-ITA, instead, are two LLMs specifically fine-tuned on Italian instructions.


\begin{table*}[htb]
  \caption{Comparing SLIMER-IT based on different backbones, with and without Definition and Guidelines (D\&G) in the prompt. LLMs with $\dag$ symbol were instruction-tuned on Italian. In parentheses the $(\pm \Delta F1)$ of performance given by the usage of D\&G.}
    \label{tab:w_wo_DeG_table}
  \centering
  \resizebox{\textwidth}{!}{
  \begin{tabular}{ccc|c|cc|c}
    \toprule
    \textbf{Backbone} & \textbf{Params} & \textbf{w/ D\&G} & \textbf{In-Domain} & \multicolumn{2}{c|}{\textbf{OOD}} & \textbf{unseen NEs} \\
    & & & WN & FIC & ADG & MN\\

    \midrule

    \multirow{2}{*}{Camoscio \dag} & \multirow{2}{*}{7B} & False & 81.80 & 82.44 & 79.01	& 32.28	\\
    & & True & 81.50 (-0.3) & 85.08 (+2.64) & 76.00 (-3.01) & 38.68 (+6.4) \\

    \midrule

    \multirow{2}{*}{LLaMA-2-chat} & \multirow{2}{*}{7B} & False & 80.69 & 80.45 & 73.81 & 32.38 \\
    & & True & 83.24 (+2.55) & 88.81 \textbf{(+8.36)} & 79.26 (+5.45) & 35.16 (+2.78) \\

    \midrule
    
    \multirow{2}{*}{Mistral-Instruct} & \multirow{2}{*}{7B} & False & 82.71 & 85.61 & 75.80	& 35.63	\\
    & & True & 85.55 \textbf{(+2.84)} & \textbf{92.78} (+7.17) & 80.56 (+4.76) & 40.64 (+5.01) \\

    \midrule
    
    \multirow{2}{*}{LLaMA-3-Instruct} & \multirow{2}{*}{8B} & False & 85.93 & 82.85 & 80.00 & 27.62	\\
    & & True & 85.38 (-0.55) & 84.38 (+1.53) & 78.29 (-1.71) & 50.74 (+23.12) \\ 
    
    \midrule
    
    \multirow{2}{*}{LLaMAntino-3-ANITA \dag} & \multirow{2}{*}{8B} & False & 84.12 & 77.06 & 74.35 & 30.90 \\
    & & True & \textbf{85.78} (+1.66) & 82.52 (+5.46) & \textbf{81.65 (+7.30)} & \textbf{54.65 (+23.75)} \\
    
    \bottomrule
  \end{tabular}
  }
\end{table*}

\subsection{Compared Models}
We compare the SLIMER-IT approach, implemented with different backbones, against other state-of-the-art approaches for zero-shot NER. All the methods are trained and evaluated in the defined zero-shot NER framework for a fair comparison. We evaluate against:

\paragraph{Token classification.} Although certainly not being suited for zero-shot NER, due to its architectural inability to cope with unseen tags, we decided to evaluate the most known approach to NER as baseline. 
As in NERMuD \cite{Aprosio2023NERMuDAE}, we use the training framework \textit{dhfbk/bert-ner}\footnote{\url{https://github.com/dhfbk/bert-ner}}. We fine-tune two different base models, \textit{bert-base-cased}, pre-trained on English, and \textit{dbmdz/bert-base-italian-cased}\footnote{\url{https://huggingface.co/dbmdz/bert-base-italian-cased}}, an Italian version.

\paragraph{GNER.} It is the best performing approach on zero-shot NER in OOD English benchmark. In GNER \cite{ding2024rethinking}, they propose a BIO-like generation, replicating in output the same input text, along with a token-by-token BIO label. Here, we consider LLaMAntino-3 as its backbone.  

\paragraph{GLiNER.} Differently from all other methods, GLiNER is based on a smaller encoder-only model, combined with a span classifier head, able to achieve competitive performance on the OOD English benchmark at a lower computational cost. We fine-tune it both using its original \textit{deberta-v3-large} English backbone and the Italian \textit{dbmdz/bert-base-italian-cased} model.

\paragraph{extremITLLaMA.} Already described in Section \ref{sec:related_works}, it represents an interesting approach to compare against. 
Based on Camoscio LLM, we compare it with SLIMER-IT approach implemented with the same backbone.

\subsection{Experimental setup}
We kept the same training configuration of SLIMER~\cite{SLIMER} on English, except that we trained on all available samples.
Depending on the backbone, the instruction-tuning prompt (see Figure~\ref{fig:SLIMER-IT prompt}) was adjusted accordingly to the structure of its template (e.g. [INST] or <|start\_header\_id|> formats).
For all the competitors, we replicated their training setup using their scripts and suggested hyper-parameters.
For the evaluation, we use the micro-F1 as computed in the UniNER\footnote{\url{https://github.com/universal-ner}} implementation.

\begin{figure}[ht]
  \centering  
  \includegraphics[width=\linewidth]{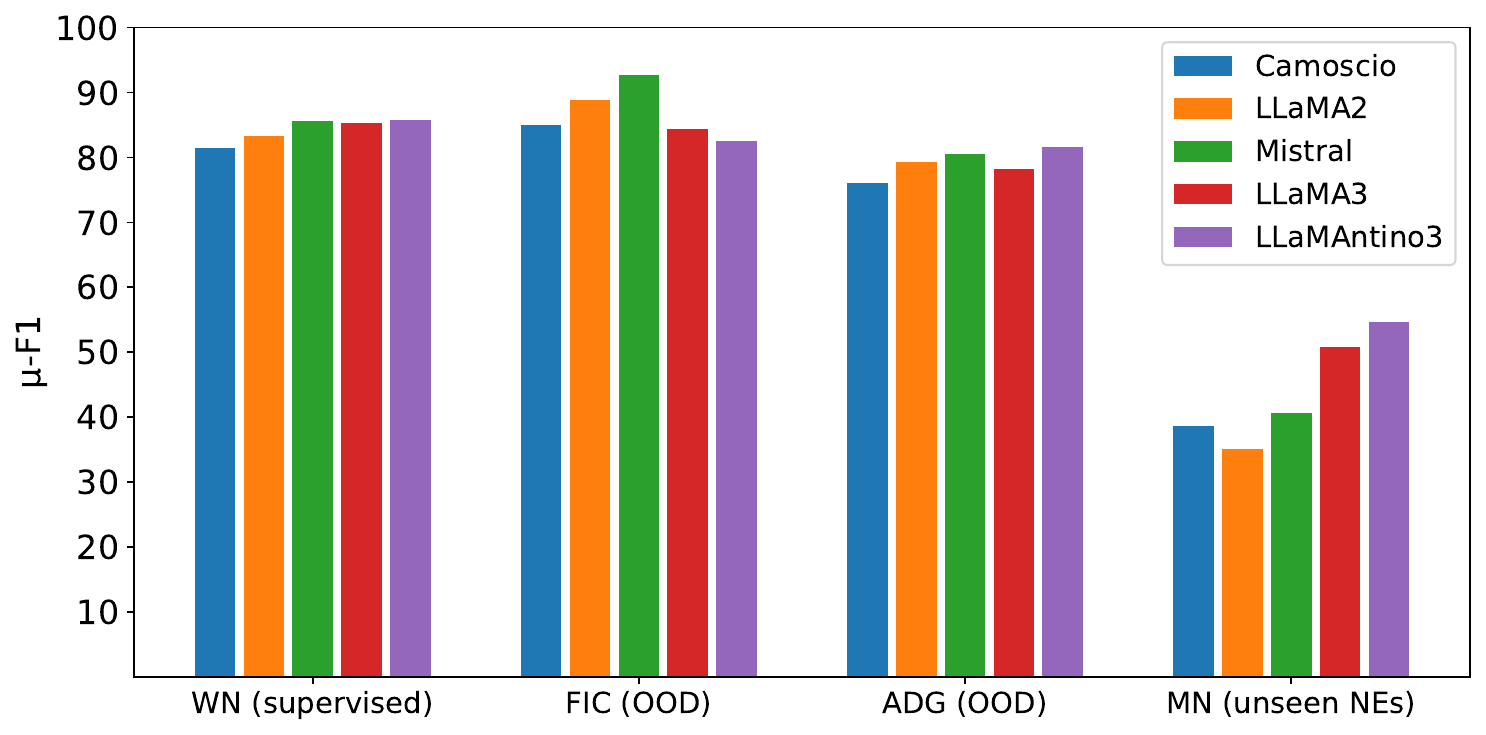}
  \caption{SLIMER-IT performance for different backbones.}
  \label{fig:comparing_SLIMER_backbones_histogram}
\end{figure}

\begin{table}[t]
    \caption{Comparison with existing off-the-shelf models for zero-shot NER on Italian. We omit in-domain evaluation to not disadvantage them against SLIMER-IT. 
    }
    \label{tab:GLiNER_models}
    \centering
    \resizebox{\linewidth}{!}{
    \begin{tabular}{l|cc|c}
    \toprule
    \textbf{Model} & \multicolumn{2}{c|}{\textbf{OOD}} & \textbf{unseen NEs} \\
    & FIC & ADG & MN \\
    \midrule

    Universal-NER-ITA & 32.4 & 43.2 & 12.8 (all seen)\\
    
    GLiNER-ITA-Large & 36.6 & 42.0 & 15.5 (all seen)\\

    GLiNER-ML & 46.5 & 49.4 & 17.4 (all seen)\\

    \midrule

    SLIMER-IT & \textbf{82.5} & \textbf{81.7} & \textbf{54.7}\\

    \bottomrule

\end{tabular}
}
\end{table}

\subsection{Results}

\paragraph{Impact of Definition and Guidelines (D\&G).} We compare SLIMER-IT with a version devoid of definition and guidelines in the prompt. To demonstrate the robustness of the approach, we train several SLIMER-IT instances, based on different LLM backbones. In Table \ref{tab:w_wo_DeG_table}, we report the results, highlighting the absolute difference in performance between the model steered by D\&Gs and the one not using them. Generally, definition and guidelines yield improvements in F1. In particular, the gap is contained when evaluating on in-domain data, whereas it becomes significant in OOD and even more substantial in unseen NEs. This is expected since D\&G help the most in conditions unseen during training. Notably, LLaMA-3-based backbones benefit the most from definition and guidelines, with improvements beyond 23 absolute F1 points, surpassing all the other models by substantial margins in never-seen-before entity tags. Some qualitative examples are shown in Appendix~\ref{app:examples}.

\paragraph{Impact of Backbones.} Regarding the choice of the SLIMER-IT backbone, we better illustrate results in Figure~\ref{fig:comparing_SLIMER_backbones_histogram}.
We can observe no remarkable difference in in-domain evaluation, where most recent models outperform older ones, as one might expect.
Also globally, Camoscio and LLaMA-2-chat obtain lower scores than the rest of the backbones, with the only exception of FIC dataset, where LLaMA-3 based architecture under-perform. However, LLaMAntino-3-ANITA reaches the best performance on 3 out of 4 datasets, with a strong gap especially in unseen named entities scenario, the most challenging one. 
Interestingly enough, thanks to their better understanding capabilities, backbones specialized on Italian are particularly effective in the unseen NEs scenario. This is the case of LLaMAntino-3-ANITA and even Camoscio, which demonstrates higher F1 than LLaMA-2. 

\paragraph{Off-the-shelf Italian NER models.} Although there has been no prior work defining a Zero-Shot NER evaluation framework for Italian, there exist fine-tune specialized state-of-the-art zero-shot NER models for Italian language.
In particular, we consider: GLiNER-ML~\cite{zaratiana2023gliner}, a multilingual instance of GLiNER, Universal-NER-ITA\footnote{\url{https://huggingface.co/DeepMount00/universal_ner_ita}} and GLiNER-ITA-Large\footnote{\url{https://huggingface.co/DeepMount00/GLiNER_ITA_LARGE}}, both specialized on Italian. These models were trained on synthetic data covering a vast number of different entity classes (up to 97k). Thus, it is impossible to directly compare them in a pure zero-shot framework, since there are no entity tags actually never-seen-before during training. However, we still report their results against SLIMER-IT. Table \ref{tab:GLiNER_models} reports the results. 
Despite this advantage, SLIMER-IT outperforms all these models by large a margin. 

\begin{table*}[ht!]
  \caption{Comparing SLIMER-IT with state-of-the-art approaches trained in the same zero-shot setting, and adopting the same backbone when possible. *Note that extremITLLaMA was fine-tuned also on the FIC and ADG train sets for the NERMuD task, so these datasets are not actually OOD for this model.}
  \label{tab:results}
  \centering
  \resizebox{\textwidth}{!}{
   \begin{tabular}{lccr|c|cc|c}
    \toprule
    \textbf{Approach} & \textbf{Backbone} & \textbf{Language} & \textbf{Params} & \textbf{In-Domain} & \multicolumn{2}{c|}{\textbf{OOD}} & \textbf{unseen NEs} \\
    & & & & WN & FIC & ADG & MN\\
    \midrule

    Token classification & BERT-base & EN & 0.11B & 83.9 & 75.6 & 75.0 & - \\
    
    Token classification & BERT-base & IT & 0.11B & 89.8 & 87.0 & 82.3 & - \\
    
    \midrule
    
    GLiNER & deberta-v3-large & EN & 0.44B & 87.8 & 77.2 & 80.3 & 0.2 \\
    
    GLiNER & BERT-base & IT & 0.11B & 89.3 & 87.5 & \textbf{84.9} & 0.6 \\
    
    \midrule
    
    extremITLLaMA & Camoscio & IT & 7B & 89.1 & 90.3* & 83.4* & 0.2 \\
    SLIMER-IT & Camoscio & IT & 7B & 81.5 & 85.1 & 76.0 & 38.7 \\
    
    \midrule
    
    GNER & LLaMAntino-3 & IT & 8B & \textbf{90.3} & \textbf{88.9} & 82.5 & 1.2 \\
    
    SLIMER-IT & LLaMAntino-3 & IT & 8B & 85.8 & 82.5 & 81.7 & \textbf{54.7} \\
    
    \bottomrule

\end{tabular}
}
\end{table*}

\paragraph{State-of-the-art comparison.} Thanks to the definition of our zero-shot evaluation framework, we can compare different state-of-the-art approaches fairly. Results are outlined in Table~\ref{tab:results}. 
When evaluating in the same domain where the model was trained, encoder-only architectures obtain strong results despite being much smaller models. This result is not surprising, given the acknowledged performance of these architectures for supervised NER. More unexpected is their ability to generalize well to OOD inputs. Also GNER proves to be quite competitive achieving the best results in in-domain evaluation, and in OOD on FIC dataset. However, all these approaches dramatically fail on never-seen-before tags, in contrast to SLIMER-IT that achieves almost 55 F1 score points. Compared with LLM-based approaches like GNER and extremITLLaMA, this proves once again that without definition and guidelines LLMs struggle in tagging novel kind of entities.

\section{Conclusions}
In this paper, we proposed an evaluation framework for Zero-Shot NER that we applied to Italian. Thanks to such a framework, we can better investigate different zero-shot properties depending on the scenario (in-domain, OOD, unseen NEs). On top of that, we compared several state-of-the-art approaches, with particular focus on SLIMER, which, thanks to the usage of definition and guidelines, is well suited to deal with novel entity types. Indeed, SLIMER-IT, our fine-tuned model based on LLaMAntino-3, surpasses other state-of-the-art techniques by large margins.
In the future, we plan to further extend the zero-shot NER benchmark, and implement an input caching mechanism for scalability to large label sets.



\clearpage

\section*{Acknowledgments}
\label{sec:funding}

The work was partially funded by:
\begin{itemize}
    \item ``ReSpiRA - REplicabilità, SPIegabilità e Ragionamento'', a project financed by FAIR, Affiliated to spoke no. 2, falling within the PNRR MUR programme, Mission 4, Component 2, Investment 1.3, D.D. No. 341 of 03/15/2022, Project PE0000013, CUP B43D22000900004  \footnote{RESPIRA: \url{https://www.opencup.gov.it/portale/web/opencup/home/progetto/-/cup/B43D22000900004}};
    \item ``MAESTRO - Mitigare le Allucinazioni dei Large Language Models: ESTRazione di informazioni Ottimizzate'' a project funded by Provincia Autonoma di Trento with the Lp 6/99 Art. 5:ricerca e sviluppo, PAT/RFS067-05/06/2024-0428372, CUP: C79J23001170001  \footnote{MAESTRO: \url{https://www.opencup.gov.it/portale/web/opencup/home/progetto/-/cup/C79J23001170001}};
    %
    %
    \item ``enRichMyData - Enabling Data Enrichment Pipelines for AI-driven Business Products and Services'', an Horizon Europe (HE) project, grant agreement ID: 101070284 \footnote{\url{https://doi.org/10.3030/101070284}}.
\end{itemize}


\bibliography{bibliography}

\newpage

\appendix

\section{SLIMER-IT on some NE tags}\label{app:examples}
In Table~\ref{tab:SLIMER_examples} we compare SLIMER-IT (LLaMAntino-based) with a version of it devoid of Definition and Guidelines (D\&G), in order to get a better insight into the usefulness of such components in zero-shot NER.
We present results for both unseen named entities (from Multinerd) and previously seen tags \textit{person}, \textit{location} and \textit{organization}, but in out-of-domain inputs (ADG and FIC datasets). 
The D\&G components improve performance by up to 37 points for unseen named entities, serving as a source of additional knowledge to the model and providing annotation directives about what should be labeled. Particularly for these named entities, the D\&G enhance precision by reducing the number of false positives the model would otherwise generate.
The performance gain provided by D\&G for known tags within out-of-domain inputs is smaller, with improvements of up to 17 points on some named entity tags. In this context, the definitions and guidelines serve more as a reasoning support than as a source of additional knowledge.

\begin{table*}[b!]
\caption{
    Some examples of definition and guidelines. Absolute F1 gains between SLIMER-IT and its version without definition and guidelines are reported. In green we highlight examples on unseen named entities, in blue examples on known tags such person, organization and location, but in Out-Of-Domain input distributions. 
  }
  \label{tab:SLIMER_examples}
  \centering
  \large
  \resizebox{\textwidth}{!}{
  \begin{tabular}{l|p{10cm}|cc|c}
    \toprule
    \textbf{NE (dataset)} & \multicolumn{1}{c|}{\textbf{Definition \& Guidelines}} & \textbf{w/o D\&G F1} & \textbf{w/ D\&G F1} & \textbf{$\Delta$} \textbf{F1}\\
    \midrule

    \rowcolor{LightGreen}
    Corpo celeste (MN) & Definizione: 'CORPO CELESTE' si riferisce a oggetti astronomici come pianeti, stelle, satelliti, costellazioni, galassie, comete e asteroidi. Linee guida: Evita di etichettare come 'corpo celeste' entità non direttamente collegate al campo dell'astronomia. Ad esempio, 'Vergine' potrebbe riferirsi anche a un segno astrologico, quindi il contesto è importante. Assicurati di non includere nomi di fenomeni non astronomici come 'alba' o 'tramonto'. Potresti incontrare ambiguità quando un termine è usato sia in campo astronomico che in contesti non astronomici, ad esempio 'aurora' che può riferirsi sia all'evento astronomico che al nome di persona. & 27.07 & 64.00 & +36.93 \\

    \midrule

    \rowcolor{LightGreen}
    Pianta (MN) & Definizione: 'PIANTA' si riferisce a organismi vegetali come alberi, arbusti, erbe e altre forme di vegetazione., Linee Guida: Quando identifichi entità 'pianta', assicurati di etichettare solo nomi di specie vegetali specifiche, come 'Fagus sylvatica', 'Suaeda vera', 'Betula pendula', evitando generici come 'alberi' o 'arbusti' se non accompagnati da una specificazione della specie. & 13.76 & 49.89 & +36.13 \\

    \midrule

    \rowcolor{LightGreen}
    Media (MN) & Definizione: 'MEDIA' si riferisce a entità come nomi di giornali, riviste, libri, album musicali, film, programmi televisivi, spettacoli teatrali e altre opere creative e di comunicazione., Linee Guida: Assicurati di etichettare solo nomi specifici di opere creative e di comunicazione, evitando generici come 'musica' o 'libro'. Presta attenzione alle ambiguità, ad esempio 'Apple' potrebbe riferirsi alla società tecnologica o ad un'opera d'arte. Escludi i nomi di artisti, autori o registi, che dovrebbero essere etichettati come 'persona', e nomi generici di strumenti musicali o generi letterari che non rappresentano opere specifiche. & 47.78 & 65.86 & +18.08 \\

    \midrule
    
    \rowcolor{LightCyan}
    Luogo (FIC) & Definizione: 'LUOGO' denota nomi propri di luoghi geografici, comprendendo città, paesi, stati, regioni, continenti, punti di interesse naturale, e indirizzi specifici., Linee Guida: Assicurati di non confondere i nomi di luoghi con nomi di persone, organizzazioni o altre entità. Ad esempio, 'Washington', potrebbe riferirsi alla città di Washington D.C. o al presidente George Washington, quindi considera attentamente il contesto. Escludi nomi di periodi storici, eventi o concetti astratti che non rappresentano luoghi fisici. Ad esempio, 'nel Rinascimento' è un periodo storico, non un luogo geografico. & 59.34 & 76.32 & +16.98 \\

    \midrule

    \rowcolor{LightCyan}
    Organizzazione (ADG) & Definizione: 'ORGANIZZAZIONE' denota nomi propri di aziende, istituzioni, gruppi o altre entità organizzative. Questo tipo di entità include sia entità private che pubbliche, come società, organizzazioni non profit, agenzie governative, università e altri gruppi strutturati. Linee Guida: Annota solo nomi propri, evita di annotare sostantivi comuni come 'azienda' o 'istituzione' a meno che non facciano parte del nome specifico dell'organizzazione. Assicurati di non annotare nomi di persone come organizzazioni, anche se contengono termini che potrebbero sembrare riferimenti a entità organizzative. Ad esempio, 'Johnson \& Johnson' è un'azienda, mentre 'Johnson' da solo potrebbe essere il cognome di una persona. & 55.56 & 71.85 & +16.29 \\

    \midrule

    \rowcolor{LightCyan}
    Persona (FIC) & Definizione: 'PERSONA' denota nomi propri di individui umani. Questo tipo di entità comprende nomi di persone reali, famose o meno, personaggi storici, e può includere anche personaggi di finzione. Linee Guida: Fai attenzione a non includere titoli o ruoli professionali senza nomi propri (es. 'il presidente' non è una 'PERSONA', ma 'il presidente Barack Obama' sì). & 79.72 & 83.33 & +3.61 \\
    
    \bottomrule
  \end{tabular}
  
}
\end{table*}

\end{document}